\documentclass{article}
\usepackage[utf8]{inputenc}
\usepackage[T1]{fontenc}

\usepackage[backend=bibtex,style=nature, url=false, doi=false, isbn=false]{biblatex}
\usepackage{amsmath}
\usepackage{amssymb}
\usepackage{amsthm}
\usepackage{microtype}
\usepackage{enumerate}
\usepackage{graphicx}
\usepackage{wrapfig}
\usepackage{xspace}
\usepackage{hyperref}
\usepackage{xcolor}
\usepackage{tikz}
\usepackage{caption}
\usepackage{listings}
\usepackage{lineno}
\usepackage{footnote}

\bibliography{bibliography}

\hypersetup{colorlinks=true}

\usetikzlibrary{spy}
\usetikzlibrary{arrows}

{\bfseries}{\itshape}
{\bfseries}{\itshape}

\makeatletter
\def\@xfootnote[#1]{%
  \protected@xdef\@thefnmark{#1}%
  \@footnotemark\@footnotetext}
\makeatother

\graphicspath{{./pictures/}}

\title{NEFI: Network Extraction From Images}
\author{M. Dirnberger, T. Kehl and  A. Neumann\\
Max Planck Institute for Informatics\\
 Saarbr\"ucken, Germany}

\begin{document}
\maketitle

\begin{abstract}	% 260 words
	Networks and network-like structures are amongst the central building blocks of many technological and biological systems.
	Given a mathematical graph representation of a network, methods from graph theory enable a precise investigation of its properties.	Software for the analysis of graphs is widely available~\cite{ICWSM09154,snap,batagelj1998pajek,5437689,loscalzo2008social,hagberg2008exploring} and has been applied to graphs describing large scale networks such as social networks, protein-interaction networks, etc.~\cite{mislove2007measurement,socialnetworks2010, Xenarios01012002, newman2003}. In these applications, graph acquisition, i.e., the extraction of a mathematical graph from a network, is relatively simple.
	However, for many network-like structures, e.g. leaf venations, slime molds and mud cracks, data collection relies on images where  graph extraction requires domain-specific solutions or even manual approaches~\cite{Price01012011, dhondt2012quantitative, baumgarten2010detection, PhysRevE.82.046118}. 

	Here we introduce \emph{Network Extraction From Images}, NEFI, a software tool that automatically extracts accurate graphs from images of a wide range of networks originating in various domains. While there is previous work on graph extraction from images, theoretical results are fully accessible only to an expert audience and ready-to-use implementations for non-experts are rarely available or insufficiently documented~\cite{obara2012bioimage, chai2013recovering, krause2013fast, nglt2006, hewrw2010}.

	NEFI provides a novel platform allowing practitioners from many disciplines to easily extract graph representations from images by supplying flexible tools from image processing, computer vision and graph theory bundled in a convenient package. 
	Thus, NEFI constitutes a scalable alternative to tedious and error-prone manual graph extraction and special purpose tools. 
	
	We anticipate NEFI to enable the collection of larger datasets by reducing the time spent on graph extraction. The analysis of these new datasets may open up the possibility to gain new insights into the structure and function of various types of networks. NEFI is open source and available at~\href{http://nefi.mpi-inf.mpg.de}{http://nefi.mpi-inf.mpg.de}.
\end{abstract}

\section{The Problem, Related Work and Motivation}		% 721 words

	The study of complex network-like objects is of increasing importance for many scientific domains. 
	The mathematical study of networks, Graph Theory, formalizes a network's structure by modeling the constituents of a network as \emph{vertices} and the pairwise relations between them as \emph{edges}\footnote{Some communities traditionally refer to vertices as nodes or sites and to edges as arcs or links.}. Networks are ubiquitous in everyday life. Examples are as diverse as the Internet, social networks, transportation networks, metabolic networks, blood vessels or the vein networks of leaves. For a comprehensive review see~\cite{newman2003}. 

	In situations where the extraction of a mathematical graph from a physical network is easy, the size of graphs that can be analyzed quickly increased from hundreds to millions of vertices. At the same time it became feasible to build large databases of various types of networks. This enabled the application of software incorporating methods from statistics and graph theory to obtain many results that changed our understanding of large scale network structures. However, digitization remains difficult for many types of networks, e.g. leaf venations, and therefore ready-to-analyze datasets are often not available. In these cases, investigation on a larger scale requires tedious and error prone data acquisition.

	In many experimental settings networks are initially available as images and it is necessary to extract the associated graphs from these images before any analysis can take place. This requires the identification of vertices and edges within the depicted structure. As this is a very work-intensive process, automated solutions are needed. 

	Leveraging advances in computer vision, several authors have proposed and successfully implemented solutions for domain specific graph extraction applications. 
	
	The authors of~\cite{obara2012bioimage,obara2012contrast} consider the mycelial networks of P.\ impudicus. They use
	watershed segmentation in combination with a novel enhancement step designed to highlight curvilinear features in the input networks.
	Based on the segmented image a skeleton is computed and used to extract the graph representing the input network.
	The resulting method is designed to be brightness and contrast invariant in order to correctly extract the networks grown by P.\ impudicus from challenging noisy or low contrast images.
	
	Baumgarten et~al.~\cite{baumgarten2010detection,baumgarten2012computational} investigate the vein networks of P.\ polycephalum. For segmenting the input image they rely on careful constant thresholding followed by a sequence of restoration algorithms. Next, the restored segmented image is used to compute a skeleton. After	applying another sequence of correction steps, the skeleton is scanned to extract the graph of the input network. The proposed approach is straight-forward and designed with a focus on images produced under controlled lab conditions.
 
	In \cite{chai2013recovering} a more general algorithm applicable for a variety of problems is proposed. Based on an original stochastic model, the authors use Monte Carlo sampling to obtain junction-points in the input image. This technically involved solution guarantees structural coherence for the resulting graph representation.
	Further examples include the extraction of road networks~\cite{nglt2006}, retinal blood vessel analysis~\cite{krause2013fast} and the extraction of plane graphs~\cite{hewrw2010}.

	The above mentioned algorithmic solutions for the network extraction problem exhibit one or more of the following
	limitations: 

	\begin{itemize}
		\item They do not build on top of well-established computer vision methods and tend to rely on ad-hoc algorithms. As a result the quality of the solution and its implementation could likely be improved. In addition, a lot of time is spent on reimplementing algorithms that are already available. 

		\item They are not implemented or only available as pseudo-code.

		\item They are implemented but not designed for easy of use, distribution and extendability.
	\end{itemize}

	We are aware that the primary objective of the work cited above is not the production of reusable software, but of tools supporting a concrete research question.
	As a result, the above authors have limited time for researching the latest advances in computer vision, software engineering techniques or writing documentation. While we understand that under these circumstances the aforementioned limitations arise naturally, we strongly believe that it is necessary to overcome those limitations in order to increase the value and the impact of network extraction software. This has become the major motivation in developing NEFI. 

	Our goal is to enable virtually anyone to automatically extract networks from images. To this end we present an extensible framework of interchangeable algorithms accessible for the non-expert through an intuitive graphical user interface. Simultaneously, we envision NEFI as a flexible platform inviting experts in computer vision, image processing and software development to improve and extend NEFI's capabilities and thus promoting their own work to a wide interdisciplinary audience of users.

\section{Network Extraction From Images}	%	430 words

	NEFI is a collection of image processing routines, segmentation methods and graph algorithms designed to process 2D digital images of various networks and network-like structures. Its main function is executing a so-called extraction pipeline, designed to analyze the structures depicted in the input image. An extraction pipeline, for short pipeline, denotes an ordered sequence of algorithms. A successful execution will return a representation of the network in terms of a weighted undirected planar graph. Computed weights include edge lengths and edge widths. Once the graph is obtained, available graph analysis software~\cite{ICWSM09154,snap,batagelj1998pajek,5437689,loscalzo2008social,hagberg2008exploring} or custom written scripts can be deployed to investigate its properties (see Supplementary Information).

\begin{figure}
			\begin{center}
			\includegraphics[width=0.48\textwidth]{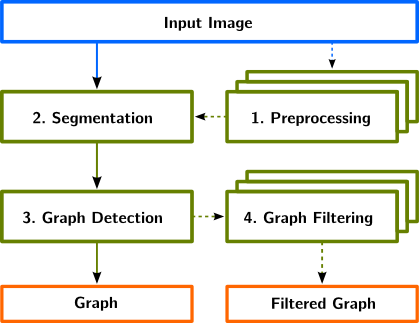}
			\end{center}
			\caption{A flow chart illustrating NEFI's pipeline components in green boxes. Dashed arrows depict optional sections of the pipeline. Blue and orange boxes denote NEFI's input and outputs respectively.}
			\label{fig::workflow}
		\end{figure}

	A typical pipeline combines algorithms from up to four different classes: preprocessing, segmentation, graph detection and graph filtering, see Figure~\ref{fig::workflow}. For each class, NEFI typically offers several interchangeable algorithms to choose from. After executing preprocessing routines, a segmentation algorithm separates foreground from background. Then the foreground is thinned to a skeleton from which the vertices and edges of the graph are determined. In the process various edge weights are computed. Finally, the graph can be subjected to a variety of useful graph filters (see Supplementary Figure~\ref{fig::sup::pipeline}).

	There are a number of predefined pipelines to get started immediately and with minimal effort. Alternatively, users may freely combine the various methods to build custom pipelines. Both approaches allow the user to experiment with the available methods in order to close in on the optimal settings for the data. Once a pipeline is constructed, it can be saved and reused. NEFI's simple pipeline concept together with a self-explanatory graphical user interface make working with NEFI intuitive and straightforward (see Supplementary Figure~\ref{fig::sup::gui}). NEFI also offers a commandline mode, which is suited for batch processing. 

\begin{figure}[t]
			\centering
			\begin{minipage}{.5\textwidth}
			  \centering
			    \includegraphics[width=0.75\textwidth]{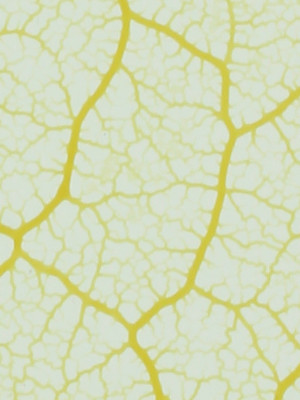}
			  % \caption{}
			  % \label{fig:neural_net}
			\end{minipage}%
			\begin{minipage}{.5\textwidth}
			  \centering
			    \includegraphics[width=0.75\textwidth]{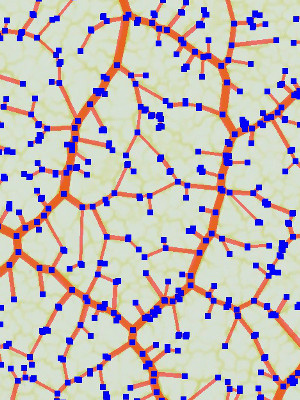}
			  % \caption{}
			  % \label{fig:function}
			\end{minipage}
			\caption{Extracted graph of the network formed by a slime mold (\emph{Physarum polycephalum}). The left hand side shows the input image depicting the network. The right hand side shows the extracted graph overlayed on top off the same image for direct comparison. The image was produced in a collaboration with the KIST Europe.}
			\label{fig::physarum}
		\end{figure}

\begin{figure}

		    	\begin{tikzpicture}

			        \begin{scope}[spy using outlines={rectangle,black,magnification=2.5,size=4cm}]

			        \node {\pgfimage[width=\textwidth]{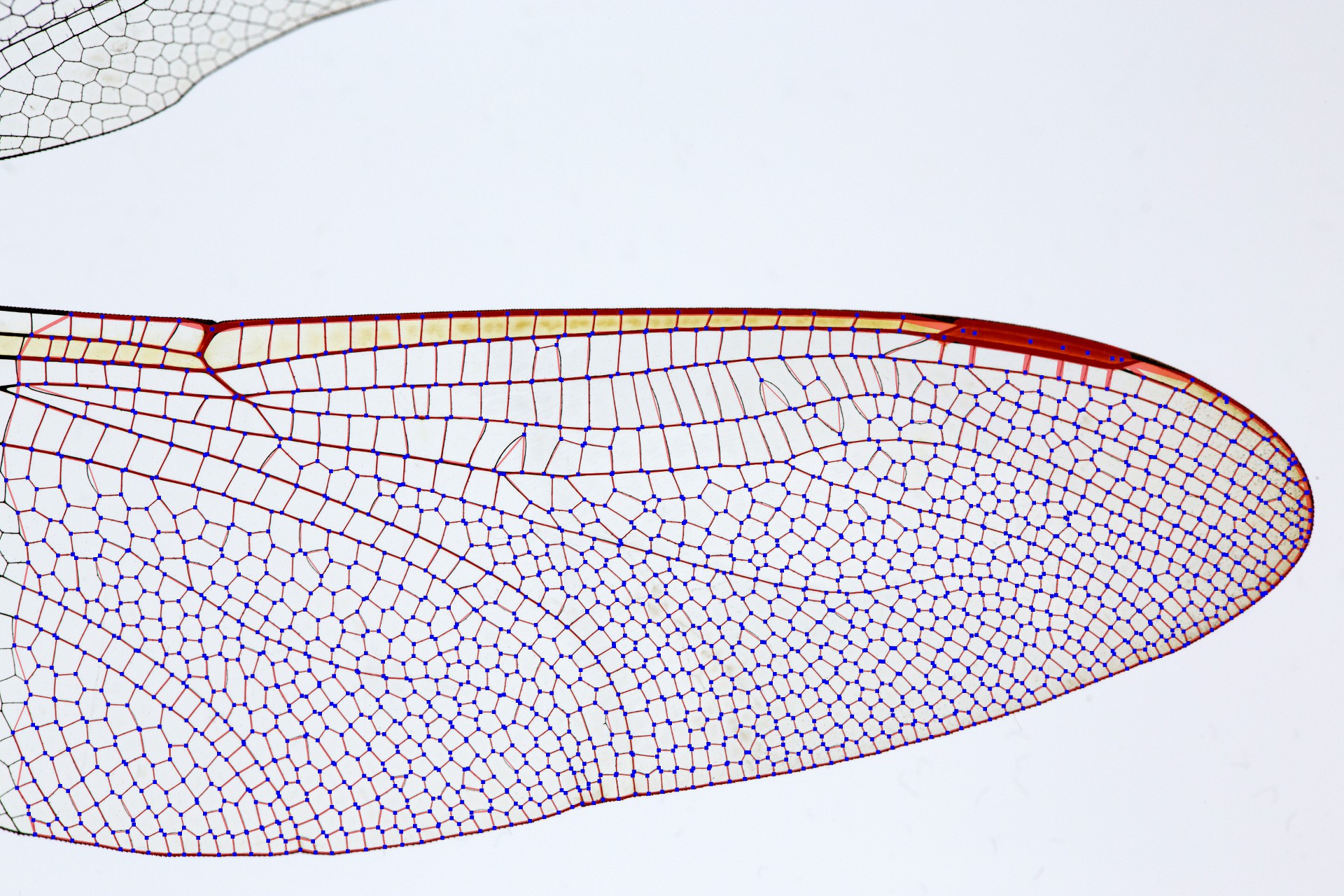}};
			        \spy on (-1,-1) in node (a) [left] at (5.5,-1.5);

			        \end{scope}

			        % \draw[very thin] (tikzspyinnode.north east) -- (tikzspyonnode.north east);
			        % \draw[very thin] (tikzspyinnode.south west) -- (tikzspyonnode.south west);

		    	\end{tikzpicture}

		    	\caption{Extracted graph of the vein network exhibited by a wing of a dragonfly (\emph{Ajax junius}). Image courtesy of Pam and Richard Winegar.}
				\label{fig::dragonlfy}

	    \end{figure}

	NEFI comes with a number of example images from different domains which we use to produce the figures in this work. Figure~\ref{fig::physarum} and Figure~\ref{fig::dragonlfy} show NEFI's output on two example images using predefined pipelines. Blue squares denote the vertices and red lines the edges of the detected graph. The thickness of the detected edges corresponds to thickness of the depicted structures. For comparison the graph is drawn on top of the input image.

	We stress that NEFI can deal with a range of inputs from various domains. In addition to the examples shown above, it has been successfully used to process images of natural (e.g. leaf venation, patterns of mud cracks) as well as man-made structures (tilings). It is also straightforward to add custom extensions. We provide a well documented framework that allows programmers to include more specialized segmentation algorithms or additional graph filters. For an overview of alternative graph extraction approaches see for example~\cite{dehkordi2011review}.

\section{Performance, Limitations and Comparison with similar Software}		%	632 words

	Depending on the input image and the pipeline, the quality of the resulting graph may vary. The major factor determining the quality of the extracted graph is the segmentation step. That is, if the image is segmented reasonably well, then the resulting graph can be expected to be accurate. In addition, graph filters can be deployed to remove stray vertices. 

	A quantitative assessment of the segmentation step is difficult as we are mainly interested in its influence on the quality of the detected graph. The problem is twofold: First, two different segmented images may lead to exactly the same graph. This becomes more likely after the application of graph filters. Second, a measure of the agreement between the depicted network and the extracted graph is not well-defined and best assessed via visual inspection. For this reason NEFI's GUI allows quick comparison of input and output (see Supplementary Figure~\ref{fig::sup::gui}).

	In contrast to other methods, we do not try to ``repair'' the segmented images or the skeleton using heuristics or user assisted methods. We thus retain a maximum of structural information which can be exploited by graph filters, allowing them to correctly remove artifacts after the graph has been established. For example, noisy regions in the input image may lead to a large number of spurious small connected components which can reliably be removed once the graph is available. Since the effects of direct manipulations on the graph itself can immediately be evaluated by visual inspection, we prefer graph filtering over less transparent or more tedious approaches that take place before the graph was found (see Supplementary Figure~\ref{fig::sup::pipeline}).

	NEFI was designed to process images produced under controlled laboratory conditions. Therefore, its general purpose methods do not work very well if the input contains irregular background or color/brightness gradients, has low contrast or insufficient resolution to detect structures that are either too dense or too fine. 
	Nevertheless, NEFI aims to compensate for some of its limitations by offering the possibility to integrate additional algorithms with comparably little effort. NEFI should be regarded as a flexible platform suitable for further development rather than a universal solution to the network extraction problem. 
	
	Due to the nature of NEFI's vertex detection, no vertices of degree two are detected. Furthermore, vertices of high degree (4 or more) are split into several degree three vertices. The latter can be merged by a suitable graph filter.

	NEFI was designed to efficiently process large quantities of images. To this end it outsources much of its
	computationally intensive tasks to highly optimized and reliable libraries such as~\href{http://opencv.org/}{OpenCV}~\cite{opencv} and~\href{https://networkx.github.io/documentation/latest/index.html}{NetworkX}~\cite{networkx}. Table~\ref{table::timings} illustrates the effectiveness of some of NEFI's algorithms. 

	To assess NEFI's value, we present a comparison with LEAF GUI~\cite{Price01012011}, which we, to the best of our knowledge, consider our closest competitor in network extraction. LEAF GUI is a specialized MATLAB application geared towards investigation of leaf veins and areoles. The tool offers a comprehensive array of functionality accessible via a well-structured GUI. Although the usability and performance of LEAF GUI is good, NEFI implements a number of important improvements: NEFI improves on LEAF GUI by offering more sophisticated segmentation algorithms, i.e. guided Watershed~\cite{watershed91} and GrabCut algorithms~\cite{grabcut2004}. Additionally, NEFI implements the more reliable Guo-Hall thinning method~\cite{guo1989parallel}, guaranteeing the connectivity of the skeleton. Since the segmentation and the resulting skeleton dominate the quality of the extracted graph, these improvements are critical. Furthermore, it becomes easier to use and compare different algorithms through the use of NEFI's flexible pipeline concept. As a result the amount of user assistance NEFI requires to operate is much reduced when compared to LEAF GUI. Thus, given a suitable pipeline, batch processing of large amounts of images of comparable quality becomes a valuable option. Finally, NEFI's source code is available for inspection and additional functionality can be added and then accessed via its streamlined GUI at any point. Table~\ref{table::features} summarizes the main results of our comparison.

	\begin{table}
		\centering
	    \begin{tabular}{| l | c | c | }
	    \hline
	    Pipeline element & Image small ($1152\times864$) & Image large ($5760 \times 3840$)\\ \hline
	    Watershed~\cite{watershed91} & < 1  & 2  \\
	    % GrabCut  & 6  & 160  \\
	    Adaptive Threshold & < 1  & 7  \\
	    Guo-Hall Thinning~\cite{guo1989parallel} &  < 1  & 12  \\
	    vertex detection & < 1  & 5  \\
	    Edge detection & < 1   & 6  \\
	    Computing edge weights & < 1  & 5 \\
	    % Full pipeline 1 & ?  & ?  \\
	    % Full pipeline 2 & ?  & ?  \\
	    \hline

	    \end{tabular}
	
	    \caption{Timings of some of NEFI's pipeline elements on images of different size. All values are in seconds. The timings were obtained on a Macbook Pro notebook equipped with a 2.4 GHz Intel i5 processor and 8 GB RAM.}
	    \label{table::timings}
	\end{table}

	\begin{savenotes}
	\begin{table}
		\centering
	    \begin{tabular}{| l | c | c | }
	    \hline
	    Features & NEFI & LEAF GUI\\ \hline
	    GUI & $\checkmark$ & $\checkmark$ \\
	    preprocessing & $\checkmark$ & $\checkmark$ \\
	    cropping & $\times$ & $\checkmark$ \\
	    segmentation & $\checkmark$ & $\checkmark$\\
	    thinning & $\checkmark$ & $\checkmark$ \\
	    graph detection & $\checkmark$ & $\checkmark$\footnote{While the documentation lists the possibility of computing the adjacency matrix which defines the graph, LEAF GUI V.1 does not allow the user to access this functionality via its GUI. We expect a future update to resolve this issue.} \\
	    graph filtering & $\checkmark$ & $\times$ \\
	    graph attributes & $\checkmark$ & $\checkmark$ \\
	    predefined pipelines & $\checkmark$ & $\times$\\
	    visual inspection of (intermediate) results & $\checkmark$ & $\checkmark$ \\
	    batch processing & $\checkmark$ & $\times$ \\
	    extensions possible & $\checkmark$ & $\times$\\
	    general purpose & $\checkmark$ & $\times$ \\
	    free combination of algorithms &  $\checkmark$ & $\times$ \\
	   	
	   	\hline
	    \end{tabular}
	    \caption{A comparison of basic features between NEFI 1.0 and LEAF GUI V.1.}
	    \label{table::features}

	\end{table}
	\end{savenotes}

\section{Conclusion}	%	100 words

	We anticipate NEFI to become a valuable tool that allows scientists from any domain to automate graph extraction from images in an intuitive fashion requiring no expert knowledge. We hope that research scientists will be able to spend more time on analyzing their data and less time on processing it. By providing a flexible platform for graph extraction, we invite experts to extend and improve NEFI in order to introduce their contributions to a wider interdisciplinary audience. In the long run we would like NEFI to further the field of network analysis by promoting the creation of new network databases.

\clearpage

\printbibliography

% \bibliographystyle{plainnat}
% \bibliography{bibliography}

\clearpage

\appendix

\section{Supplementary information}

	\subsection{General Information}

		NEFI is an open source Python application and available at~\href{http://nefi.mpi-inf.mpg.de}{http://nefi.mpi-inf.mpg.de}. NEFI's homepage includes a gallery of various use-cases and a comprehensive guide containing instructions on how to download, install and use the latest version of NEFI on Windows, Mac and Linux. 

	\subsection{Pipeline and Graphical User Interface}

\begin{figure}
			\centering
			\begin{minipage}{.175\textwidth}
			  \centering
			    \includegraphics[width=0.75\textwidth]{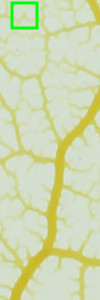}
			  % \caption{}
			  % \label{fig:neural_net}
			\end{minipage}%
			\begin{minipage}{.175\textwidth}
			  \centering
			    \includegraphics[width=0.75\textwidth]{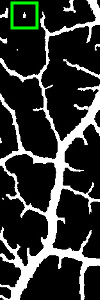}
			  % \caption{}
			  % \label{fig:neural_net}
			\end{minipage}%
			\begin{minipage}{.175\textwidth}
			  \centering
			    \includegraphics[width=0.75\textwidth]{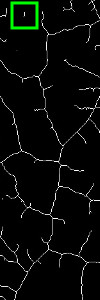}
			  % \caption{}
			  % \label{fig:function}
			\end{minipage}
			\begin{minipage}{.175\textwidth}
			  \centering
			    \includegraphics[width=0.75\textwidth]{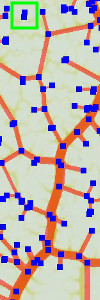}
			  % \caption{}
			  % \label{fig:neural_net}
			\end{minipage}%
			\begin{minipage}{.175\textwidth}
			  \centering
			    \includegraphics[width=0.75\textwidth]{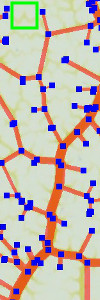}
			  % \caption{}
			  % \label{fig:neural_net}
			\end{minipage}%
			\caption{Direct comparison of NEFI's pipeline steps given a slice of an image of a slime mold (\emph{Physarum polycephalum}). From left to right: input image, segmented image, skeletonized image, detected graph and filtered graph. The green square contains a very faint vein which the segmentation does not pick up correctly. Thus, the skeleton is becomes fragmented which leads to spurious vertices in the detected graph. By applying a graph filter we remove unwanted vertices without manipulation of the segmented or the skeletonized image. Similar filtering can remove "dead-ends", i.e. vertices that do not belong to any cycle in the graph.}
			\label{fig::sup::pipeline}
		\end{figure}

\begin{figure}
			\begin{center}
			\includegraphics[width=\textwidth]{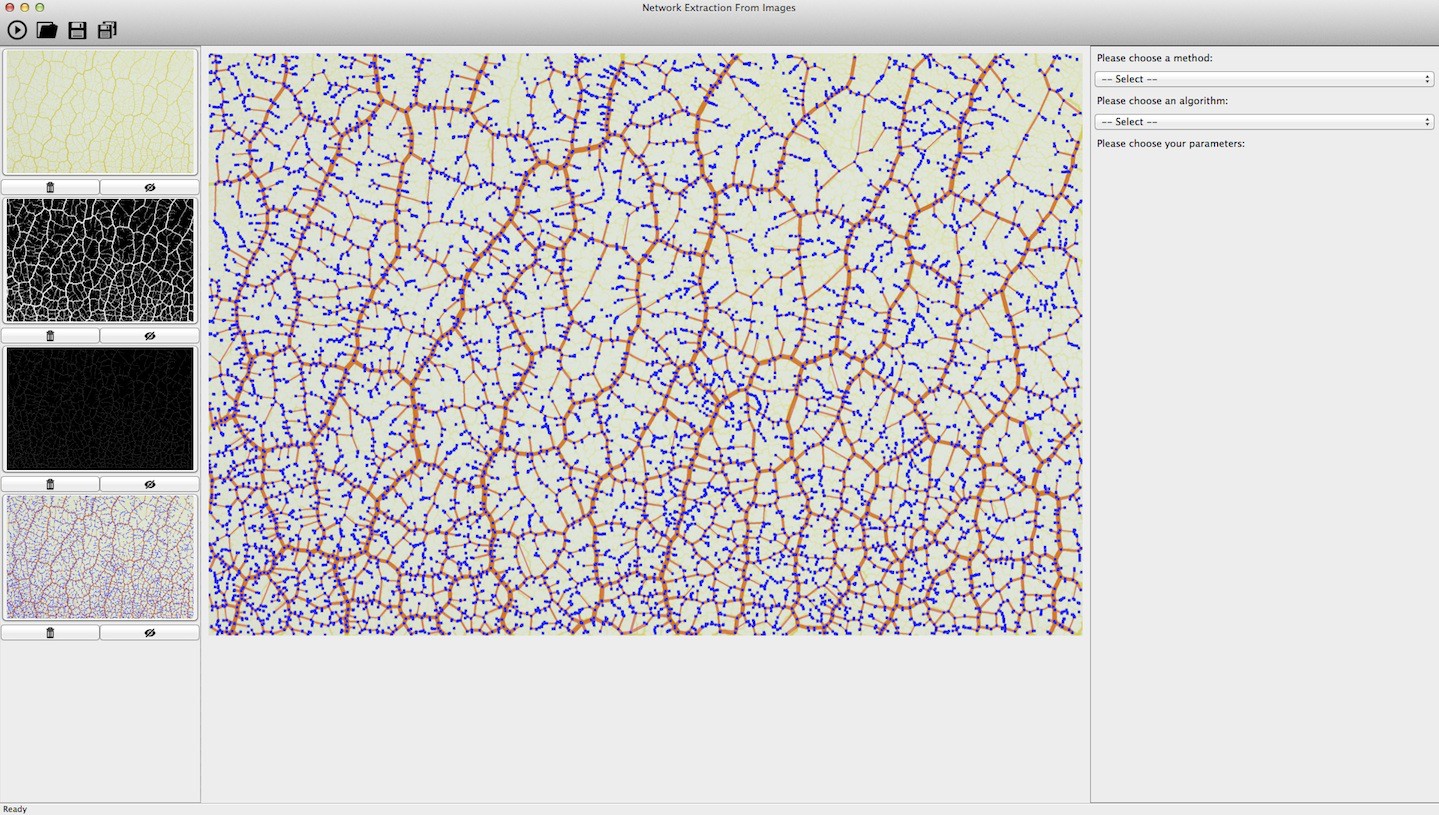}
			\end{center}
			\caption{A screenshot of NEFI's GUI running on Mac OS. On the left hand side NEFI lists intermediate results as thumbnails. Bringing the final result to the center workspace allows for direct visual assessment of the quality of the extracted graph. On the right hand side NEFI's pipeline elements can be accessed.}
			\label{fig::sup::gui}
		\end{figure}

		Figure~\ref{fig::sup::pipeline} shows the intermediate results of NEFI's pipeline listed in the order of their execution.
		When a pipeline is executed, NEFI makes all intermediate results available via its clean and intuitive GUI, see Figure~\ref{fig::sup::gui}. Using the GUI all basic functions of NEFI can be accessed in an intuitive fashion

	\subsection{Comparison between NEFI and LEAF GUI}

		To assess the difference in quality of the respective output of NEFI and LEAF GUI one would like to compare the output graphs. However, since LEAF GUI currently does not make a graph representation available, we resort to a comparison of the segmentation and the thinning steps. This is justified, since these steps determine the quality of the extracted graphs.

\begin{figure}[t]
			\centering
			\begin{minipage}{.5\textwidth}
			  \centering
			    \includegraphics[width=0.75\textwidth]{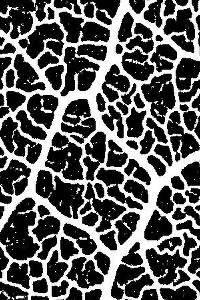}
			  \caption*{NEFI}
			  % \label{fig:neural_net}
			\end{minipage}%
			\begin{minipage}{.5\textwidth}
			  \centering
			    \includegraphics[width=0.75\textwidth]{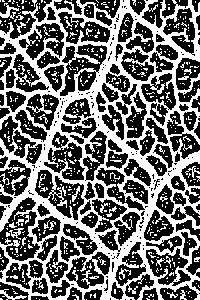}
			  \caption*{LEAF GUI}
			  % \label{fig:function}
			\end{minipage}

			\caption{Comparison of segmented images of the network formed by a slime mold (\emph{Physarum polycephalum}). Note how NEFI's fully automatic Watershed based segmentation is sensitive and robust to noise at the same time. Both results may be improved further by careful manual tuning of the algorithm settings.}
			\label{fig::sup::seg::phys}
		\end{figure}

\begin{figure}[t]
			\centering
			\begin{minipage}{.5\textwidth}
			  \centering
			    \includegraphics[width=0.75\textwidth]{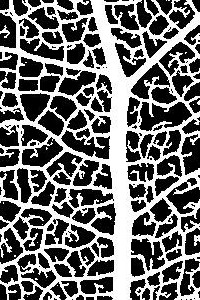}
			  % \caption{}
			  % \label{fig:neural_net}
			\end{minipage}%
			\begin{minipage}{.5\textwidth}
			  \centering
			    \includegraphics[width=0.75\textwidth]{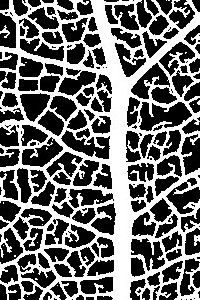}
			  % \caption{}
			  % \label{fig:function}
			\end{minipage}

			\caption{Comparison of segmented images of the vein network formed by a leaf (species unknown to the authors)~\cite{leaf}. The results are identical because the optimal segmentation algorithm was found to be adaptive thresholding, a method available both in NEFI as well as in LEAF GUI. The result indicates that LEAF GUI's segmentation is competitive within its domain of use.}
			\label{fig::sup::seg::leaf}
		\end{figure}

		For segmentation, LEAF GUI offers basic adaptive and constant thresholding. On top of LEAF GUI's algorithms, NEFI's segmentation additionally includes different variations of more advanced segmentation methods such as Watershed~\cite{watershed91} or GrabCut~\cite{grabcut2004} methods. Critically, for a general purpose tool, these allow for a wider range of images to be segmented correctly. Figure~\ref{fig::sup::seg::phys} and Figure~\ref{fig::sup::seg::leaf} compare segmentation performance on two representative images.

\begin{figure}[t]
			\centering
			\begin{minipage}{.5\textwidth}
			  \centering
			    \includegraphics[width=0.75\textwidth]{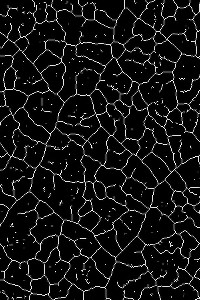}
			  \caption*{NEFI}
			  % \label{fig:neural_net}
			\end{minipage}%
			\begin{minipage}{.5\textwidth}
			  \centering
			    \includegraphics[width=0.75\textwidth]{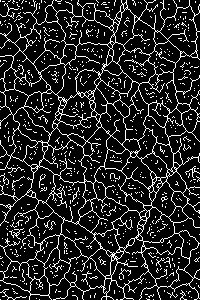}
			  \caption*{LEAF GUI}
			  % \label{fig:function}
			\end{minipage}

			\caption{Comparison of skeletonized images of the network formed by a slime mold (\emph{Physarum polycephalum}). Note how NEFI's Guo Hall thinning is much less likely to produce artifacts. LEAF GUI offers the possibility of repairing the skeleton manually.}
			\label{fig::sup::skel::phys}
		\end{figure}

\begin{figure}[t]
			\centering
			\begin{minipage}{.5\textwidth}
			  \centering
			    \includegraphics[width=0.75\textwidth]{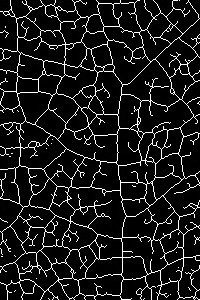}
			  \caption*{NEFI}
			  % \label{fig:neural_net}
			\end{minipage}%
			\begin{minipage}{.5\textwidth}
			  \centering
			    \includegraphics[width=0.75\textwidth]{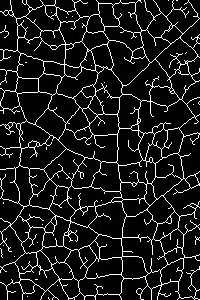}
			  \caption*{LEAF GUI}
			  % \label{fig:function}
			\end{minipage}

			\caption{Comparison of segmented images of the vein network formed by a leaf (species unknown to the authors)~\cite{leaf}. Both thinning algorithms produce results of comparable quality.}
			\label{fig::sup::skel::leaf}
		\end{figure}

		For skeletonization, LEAF GUI uses an iterative thinning approach which tends to produce a highly fragmented skeleton. In contrast, NEFI utilizes the well-established method by Guo and Hall which preserves connectivity~\cite{guo1989parallel}. In many cases this algorithm produces a superior skeleton, allowing for robust graph detection. Figure~\ref{fig::sup::skel::phys} and Figure~\ref{fig::sup::skel::leaf} show the skeleton images derived from the respective segmented images seen in Figures~\ref{fig::sup::seg::phys} and \ref{fig::sup::seg::leaf}.

	\subsection{Analysis of Graphs}

		NEFI is a tool that facilitates data acquisition, which is a necessary precursor to data analysis. While NEFI provides users with 
		valuable data in form of graphs, understanding this data is even more important. To analyze NEFI's output one can either rely on open source graph analysis software~\cite{ICWSM09154,snap,batagelj1998pajek,5437689,loscalzo2008social,hagberg2008exploring} or write custom analysis programs dealing with special situations and computing non-standard observables. The latter requires some familiarity with software libraries such as Boost in {C++} or NetworkX in Python~\cite{networkx}, which are capable of dealing with graphs. To get the user started immediately, we provide a minimal Python program that illustrates the basic steps required to perform graph analysis. The code shows how to read a graph from disk given NEFI's output and how to compute a histogram of a given edge attribute, e.g. edge length in pixel. The code can be downloaded from NEFI's project page.

% % 0260

% % 0721
% % 0430
% % 0632
% % 0100
% % 1983 

\end{document}